\pdfoutput=1
\documentclass[letterpaper]{article} 
\usepackage{aaai24}  
\usepackage{times}  
\usepackage{helvet}  
\usepackage{courier}  
\usepackage[hyphens]{url}  
\usepackage{graphicx} 
\usepackage{natbib}  
\usepackage{caption} 
\usepackage{algorithm}
\usepackage{algorithmic}
\usepackage{bm}
\usepackage{pdfpages}
\usepackage{multirow}
\usepackage{cite}
\usepackage{amsmath}

\usepackage{newfloat}
\usepackage{listings}
\DeclareCaptionStyle{ruled}{labelfont=normalfont,labelsep=colon,strut=off} 
\floatstyle{ruled}
\newfloat{listing}{tb}{lst}{}
\floatname{listing}{Listing}
\title{MFA-Net: Multi-Scale feature fusion attention network for liver tumor segmentation}
\author{
    Yanli~Yuan \textsuperscript{\rm 1},
    Bingbing~Wang \textsuperscript{\rm 1},
    Chuan~Zhang\textsuperscript{\rm 1}\thanks{Corresponding author.},
    Jingyi~Xu\textsuperscript{\rm 2},
    Ximeng~Liu\textsuperscript{\rm 3},
    Liehuang~Zhu\textsuperscript{\rm 1}
}
\affiliations{
    \textsuperscript{\rm 1}Beijing Institute of Technology \\
    \textsuperscript{\rm 2}Singapore University of Technology and Design\\
    \textsuperscript{\rm 3}Fuzhou University \\
    \{yanliyuan, bingbwang, chuanz, liehuangz\}@bit.edu.cn, jingyi\_xu@mymail.sutd.edu.sg, snbnix@gmail.com
}

\usepackage{bibentry}

\begin{document}

\maketitle

\begin{abstract}
Segmentation of organs of interest in medical CT images is beneficial for diagnosis of diseases. Though recent methods based on Fully Convolutional Neural Networks (F-CNNs) have shown success in many segmentation tasks, fusing features from images with different scales is still a challenge: (1) Due to the lack of spatial awareness, F-CNNs share the same weights at different spatial locations. (2) F-CNNs can only obtain surrounding information through local receptive fields. To address the above challenge, we propose a new segmentation framework based on attention mechanisms, named MFA-Net (\textbf{M}ulti-Scale Feature \textbf{F}usion \textbf{A}ttention \textbf{Net}work). The proposed framework can learn more meaningful feature maps among multiple scales and result in more accurate automatic segmentation. We compare our proposed MFA-Net with SOTA methods on two 2D liver CT datasets. The experimental results show that our MFA-Net produces more precise segmentation on images with different scales.
\end{abstract}

\section{Introduction}
Image segmentation is a task to divide an image into several regions and present targets of interest, which has a wide range of application scenarios \cite{yuheng2017image,liu2019recent}. Particularly, the use of image segmentation techniques in medical fields plays an important role in the diagnosis and treatment process \cite{ramesh2021review}. Now the deep learning methods are commonly used, but although much progress has been made by applying deep learning models, automatic medical image segmentation still remains a challenging task due to several difficulties, such as low image quality or biased datasets.  In order to solve these problems, the fully convolutional neural networks (F-CNNs) were proposed \cite{long2015fully}, which started the research on image segmentation at the semantic level. But many existing F-CNNs still have inherent flaws. They may lack spatial awareness and they acquire surrounding information only through local receptive fields\cite{zeng2019sese}. To further improve the performance of F-CNNs, attention mechanisms are introduced to optimize the structure of models \cite{woo2018cbam,zhang2020relation,roy2018concurrent}. Among them is SE block, which can capture channel-wise dependencies in a feature map. The advantage of SE block is achieving channel-wise recalibration without significantly increasing consumption and it can be easily applied to both shallow and deep layers.

Inspired by those previous works, we proposed a multi-scale feature fusion attention network, which we named MFA-Net, for liver tumor segmentation. The attention mechanism of our model, which we called SCSE, involves two sub modules designed based on the SE module: (1) a SSCE sub module that squeezes spatially and excites in channel, and (2) a CSSE sub module that squeezes in channel and excites spatially. SSCE can assign large weights to valid feature channels and CSSE plays a role in spatial channels. Hence, our attention mechanism SCSE is able to recalibrate the feature maps along channel and space simultaneously. 

Our major contributions are summarized as follows:

\begin{figure*}
\centering
\includegraphics[width=0.8 \textwidth]{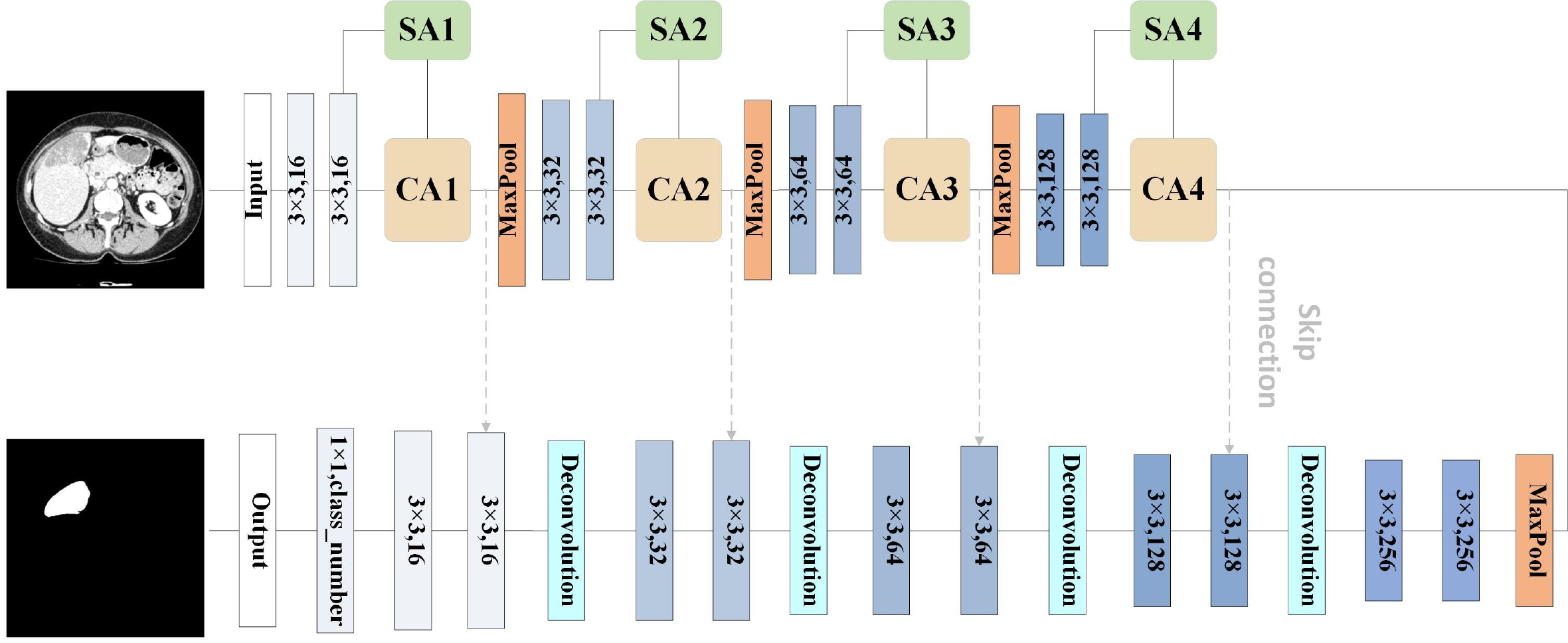}
\caption{Our proposed model MFA-Net. We adopt four channel attention blocks (${CA_{1-4}}$) and four spatial attention blocks (${SA_{1-4}}$) in the U-Net architecture. All blocks are located after every two successive convolution layers of the encoder, and the spatial attention block and the channel attention block are used in parallel each time. } \label{fig2}
\end{figure*}

\begin{itemize}
    \item We propose two new attention modules named SSCE and CSSE as two sub modules of SCSE, which is used in our model. The two sub modules are improved based on the original SE block. Compared with some existing F-CNNs, our model applies the attention mechanism to both channel dimension and spatial dimension.
    \item We propose a new model called MFA-Net by incorporating SCSE module with U-Net model. In this process, we chose to use two sub modules in parallel. Compared with serial structure, this method enables the two sub modules to function without interference from each other.
    \item We validate the performance of the proposed MFA-Net model on two liver tumor datasets, the 3D-IRCADb-01 database and the LiTS 2017 dataset. The experimental results demonstrate its effectiveness and the results are comparable to those obtained by frameworks introducing other attention mechanism modules.
\end{itemize}

\section{Related Works}
\subsection{CNNs for Image Segmentation}
The transition from CNNs to F-CNNs marks a significant advancement in the field of image segmentation. CNNs are primarily designed for classification tasks but F-CNNs have been specifically developed to address the challenges of pixel-level segmentation. F-CNNs eliminate the fully connected layers and instead employ convolutional layers throughout the entire network. The use of skip connections and upsampling techniques further enhances the spatial resolution of the segmentation results. To further solve the 3D medical image segmentation problem, a series of models such as DeepMedic\cite{kamnitsas2016deepmedic} and V-Net\cite{milletari2016v} have been proposed. Such architectures achieved excellent results in the field of medical image segmentation.

\subsection{Attention Mechanism}
To improve the feature extraction ability of the model, attention mechanism was introduced into F-CNNs. The SE block we chose as the basis of our new module is a typical attention mechanism often used in medical image segmentation. SeResUNet \cite{wen2021squeeze} was proposed to model the long-range dependencies among different channels of the learned feature maps. An architecture based on U-Net with SE block was proposed for brain tumor segmentation \cite{iantsen2021squeeze}. A few of these tasks also focused on the spatial dimension. However, it is necessary to pay attention to spatial dimension since it can help the model better understand the relationship between adjacent pixels and the utilization features. 

\section{Feature Fusion Attention Network}
In this section, we propose MFA-Net based on U-Net architecture, in which we incorporate our integration strategy and relevant blocks. In Fig. \ref{fig2}, we illustrate the MFA-Net that utilizes various attention blocks to achieve comprehensive attention guidance for both spatial and channel dimensions.

\begin{figure}
\centering
\includegraphics[width=1\linewidth]{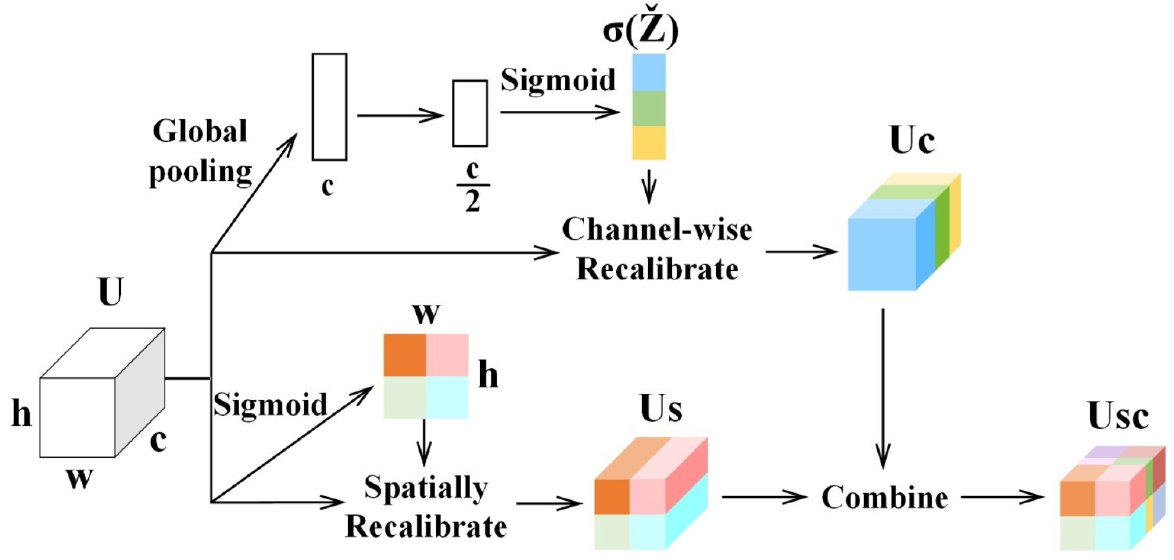}
\caption{Network architecture SCSE with squeeze \& excitation (SE) blocks.} \label{fig3}
\end{figure}

\begin{figure*}[!t]
\centering
\includegraphics[width=0.8\linewidth]{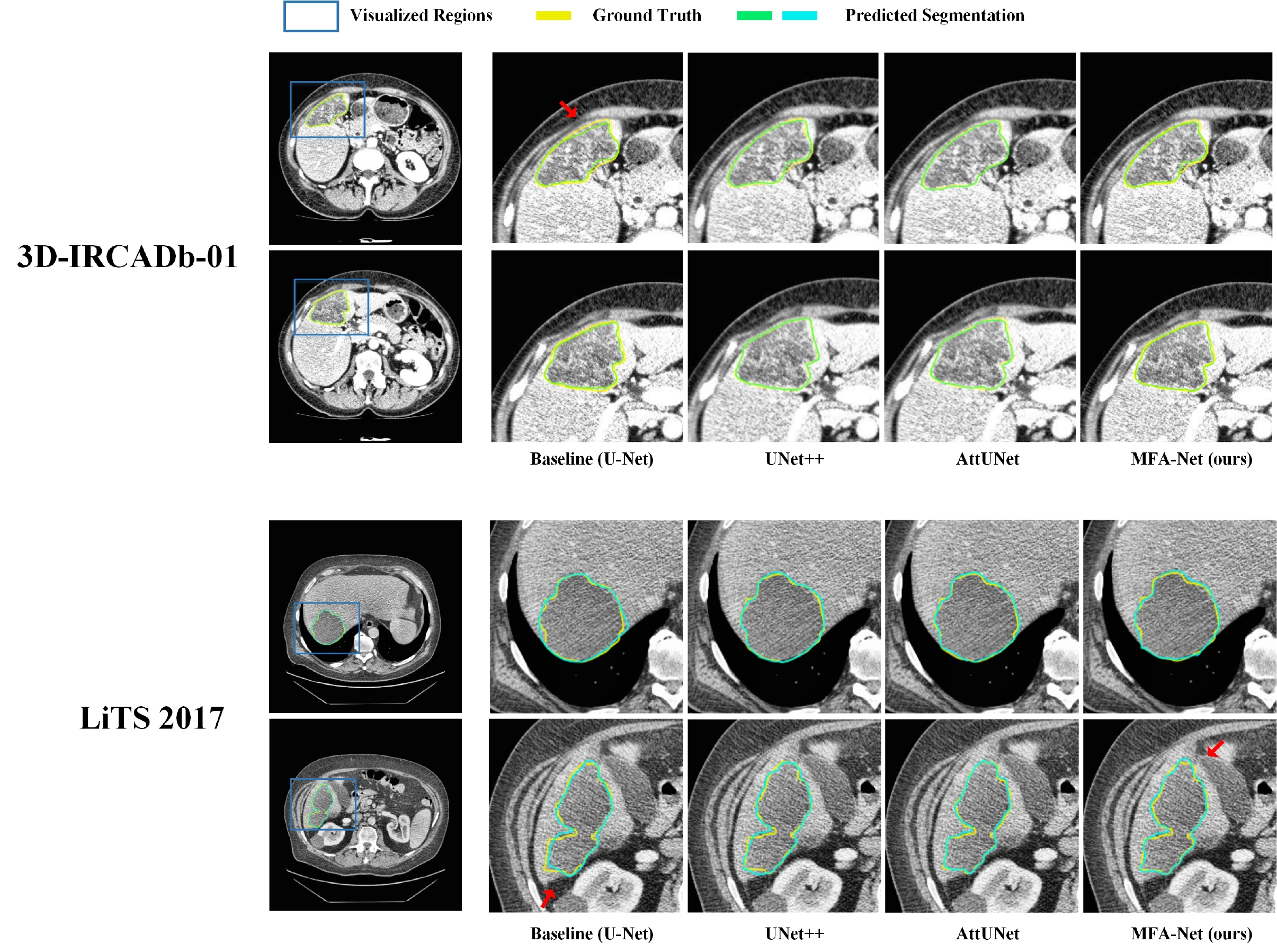}
\caption{Visual comparison between MFA-Net and other networks based on U-Net for liver tumor segmentation. Red arrows highlight some mis-segmentation. Our MFA-Net achieves better results.} \label{fig4}
\end{figure*}

\subsection{Model Architecture}

The proposed MFA-Net making use of comprehensive attention blocks is shown in Fig. \ref{fig2}, where we add specialized attention blocks to achieve comprehensive attention guidance with respect to the space and channel of the feature maps simultaneously. As shown in Fig. \ref{fig2}, our MFA-Net includes four spatial attention modules (${SA_{1-4}}$) and four channel attention modules (${CA_{1-4}}$). We use SE block in channel dimension to weight more relevant channels. For spatial dimension, we also introduce the SE block to enhance the area of interest on the feature maps. We believe that the spatial dimension is more important than the channel dimension. The attention blocks are detailed in the following.

\subsection{Modules Based On SE Block}
On the basis of the original SE block, we made some improvements and proposed three variants of SE module, which are called SSCE and CSSE, SCSE is the parallel combination of the former two. SCSE is also an important part of our MFA-Net. The following is an introduction of the two sub modules SSCE and CSSE.

\subsubsection{Spatial Squeeze and Channel Excitation (SSCE)}
In this architecture, the input feature map $\boldsymbol{U} =  [\boldsymbol{u_1},\boldsymbol{u_2},\dots,\boldsymbol{u_c}]$ is considered as a combination of channels $\boldsymbol{u_{i}}$, $\boldsymbol{u_{j}}$ $\in$ $R^{h \times w \times 1}$. The feature map $\boldsymbol{U}$ $\in$ $R^{h \times w \times c}$.

\begin{enumerate}
  \item Global average pooling. Through a global average pooling layer, the spatial squeeze is completed. After this operation, vector $\boldsymbol{Z}$ with its elements is obtained, which contains the global spatial.

  \item Convolution. For $\boldsymbol{z_k}$ $\in$ $R^{1 \times 1 \times c}$, after passing the first fully connected layer, which has a weight of $\boldsymbol{W_1}$ $\in$ $R^{c \times {\frac{c}{2}}}$, a tensor $\in$ $R^{1 \times 1 \times {\frac{c}{2}}}$ is obtained. It then passes through a nonlinear activation function ReLU $\sigma$(·) \cite{nair2010rectified}, and the resulting output passes through the second fully connected layer. After these steps, $\boldsymbol{\hat{Z}}$ is transformed from vector $\boldsymbol{z}$, and 
the size of $\boldsymbol{\hat{Z}}$ return to $\boldsymbol{\hat{Z}}$ $\in$ $R^{1 \times 1 \times c}$. 
  \item Normalization. After a sigmoid layer $\delta(\boldsymbol{\hat{z}})$, each element of $\boldsymbol{\hat{Z}}$ normalized to the range [0,1], and the value of each element is expressed as $\delta(\boldsymbol{\hat{z_k}})$. 
  \item Recalibration. The vector gained is used to recalibrate or excite $\boldsymbol{U}$  to $\boldsymbol{U_c}$. In more detail, each resulting $\delta(\boldsymbol{\hat{z_k}})$ is multiplied separately to each corresponding value of the original feature graph $\boldsymbol{U}$ to get the output $\boldsymbol{U_c}$.
\end{enumerate}

\begin{eqnarray}
		\boldsymbol{U_c} = [ \delta({\hat{z_1}})\boldsymbol{u_1},\delta({\hat{z_2}})\boldsymbol{u_2},\dots,\delta({\hat{z_c}})\boldsymbol{u_c}].
\end{eqnarray}

\begin{table*}[]
\centering
\caption{Quantitative evaluation of different SE modules on
3D-IRCADb-01 and LiTS 2017 datasets. ↑ denotes the improvement compared to the baseline method, bold results show the best scores.}\label{tab1}
\begin{tabular}{ccccccc}
\hline
\multirow{2}{*}{Network}                                       & \multicolumn{3}{c}{3D-IRCADb-01}              & \multicolumn{3}{c}{LiTS 2017}                 \\ \cline{2-7} 
 &
  \begin{tabular}[c]{@{}c@{}}Dice\\ Score (\%)\end{tabular} &
  \begin{tabular}[c]{@{}c@{}}Jaccard\\ Index (\%)\end{tabular} &
  \begin{tabular}[c]{@{}c@{}}Pixel\\ Accuracy (\%)\end{tabular} &
  \begin{tabular}[c]{@{}c@{}}Dice\\ Score (\%)\end{tabular} &
  \begin{tabular}[c]{@{}c@{}}Jaccard\\ Index (\%)\end{tabular} &
  \begin{tabular}[c]{@{}c@{}}Pixel\\ Accuracy (\%)\end{tabular} \\ \hline
Baseline (U-Net)                                               & 76.4          & 88.7          & 76.8          & 65.3          & 70.2          & 66.1          \\
U-Net++                                                        & 76.8          & 88.5          & 80.3          & 65.6          & 71.0          & 67.4          \\
AttUNet                                                        & 76.6          & 88.8          & 85.6          & \textbf{67.3} & 71.8          & \textbf{69.4}         \\
\begin{tabular}[c]{@{}c@{}}\textbf{Our Method}\end{tabular} & \textbf{77.1↑} & \textbf{90.1↑} & \textbf{86.7↑} & 67.1↑          & \textbf{71.9↑} & 69.1↑ \\ \hline
\end{tabular}
\end{table*}

\begin{table*}[]
\centering
\caption{Ablation experiments on 3D-IRCADb-01 and LiTS 2017 datasets. ‘SSSE’ and ‘CSSE’ are two attention blocks based on SE. ‘SCSE’ denotes the combination module of two blocks, which is the main module of our MFA-Net. The best results are boldfaced.}\label{tab2}
\begin{tabular}{ccccccc}
\hline
\multirow{2}{*}{Network} & \multicolumn{3}{c}{3D-IRCADb-01}              & \multicolumn{3}{c}{LiTS 2017}                 \\ \cline{2-7} 
 &
  \begin{tabular}[c]{@{}c@{}}Dice\\ Score (\%)\end{tabular} &
  \begin{tabular}[c]{@{}c@{}}Jaccard\\ Index (\%)\end{tabular} &
  \begin{tabular}[c]{@{}c@{}}Pixel\\ Accuracy (\%)\end{tabular} &
  \begin{tabular}[c]{@{}c@{}}Dice\\ Score (\%)\end{tabular} &
  \begin{tabular}[c]{@{}c@{}}Jaccard\\ Index (\%)\end{tabular} &
  \begin{tabular}[c]{@{}c@{}}Pixel\\ Accuracy (\%)\end{tabular} \\ \hline
Baseline (U-Net)         & 76.4          & 88.7          & 76.8          & 65.3          & 70.2          & 66.1          \\
Baseline+SSSE            & 75.7          & 88.9          & 83.4          & 65.6          & 71.1          & 67.7          \\
Baseline+CSSE            & 76.6          & 89.6          & 85.6          & 66.7          & 71.5          & 68.4          \\

\begin{tabular}[c]{@{}c@{}}\textbf{Baseline+SCSE}\end{tabular} & \textbf{77.1} & \textbf{90.1} & \textbf{86.7} & \textbf{67.1}          & \textbf{71.9} & \textbf{69.1} \\ \hline
\end{tabular}
\end{table*}

\subsubsection{Channel Squeeze and Spatial Excitation (CSSE)}
Here, the input feature map $\boldsymbol{U}$ is considered as $\boldsymbol{U}$ = [$\boldsymbol{u^{1,1}}$, $\boldsymbol{u^{1,2}}$, \dots, $\boldsymbol{u^{i,j}}$, \dots, $\boldsymbol{u^{h,w}}$], where $\boldsymbol{u^{i,j}}$ $\in$ $R^{1 \times 1 \times c}$ represents to the spatial location($i$,$j$) with $i$ $\in$ {1,2,$\dots$,$h$}, $j$ $\in$ {1,2,$\dots$,$w$}.

\begin{enumerate}
    \item Convolution. After a convolution operation with $\boldsymbol{W_{sq}}$ $\in$ $R^{1 \times 1 \times C \times 1}$, a new feature map $\boldsymbol{q}$ $\in$ $R^{h \times w}$ is obtained. 
    \item Normalization. New feature map $\boldsymbol{q}$ is normalized through a sigmoid layer to transform to the spatial attention map, whose value is rescaled to [0,1].
    \item Recalibration. The spatial attention map can be directly applied to the original feature map $\boldsymbol{U}$ to complete the spatial information calibration, recalibrating or exciting $\boldsymbol{U}$ to $\boldsymbol{U_s}$.
\end{enumerate}

\begin{align}
\boldsymbol{U_s} &= [ \delta(q_{1,1})\boldsymbol{u^{1,1}},\delta(q_{1,2})\boldsymbol{u^{1,2}},\dots,\nonumber\\
&\quad\quad\quad\dots,\delta(q_{i,j})\boldsymbol{u^{i,j}},\dots,\delta(q_{h,w})\boldsymbol{u^{h,w}}].
\end{align}

\subsubsection{Concurrent Spatial and Channel Squeeze and Channel Excitation (SCSE)}
SCSE is the combination of the above two sub modules. It can recalibrate the input $\boldsymbol{U}$ spatially and channel-wise simultaneously. After element-wise addition of the channel and spatial excitation, we can obtain a more accurate feature map $\boldsymbol{U_{sc}} = \boldsymbol{U_c} + \boldsymbol{U_s}$. It should be emphasized that we chose to use the two sub modules in parallel, rather than in series. This is because the feature map will be changed after processing by any sub module, so we chose a parallel mode to prevent the two sub modules from interfering with each other. When a location of the feature map $\boldsymbol{U}$ gets high attention both from channel re-scaling and spatial re-scaling, it will get higher activation. The structure of this combination module is shown as Fig. \ref{fig3}.

\section{Experiments}
In this section, we conducted lots of experiments to explore the effect of our model. We choose three classical models for comparison. For both applications, we also implemented ablation studies of our attention mechanism module.

\subsection{Experimental Setup}

The dataset used in the experiments comes from 3D-IRCADb-01 database \cite{soler20103d} and the ISBI LiTS 2017 Challenge. All the raw data is enhanced by CT-Windowing method and contrast limited adaptive histogram equalization. Our network is implemented on a single Nvidia GPU Telsa T4 (16G). Experiments are conducted using four blocks, U-Net, U-Net with only SSCE, U-Net with only CSSE, and U-Net with SCSE, respectively. In order to evaluate the proposed method, we utilize three commonly used metrics: Dice Score, Jaccard Index, and Pixel Accuracy.

\subsection{Experimental Results}
In this section, we perform a series of comparative experiments to show the effectiveness of our proposed method. The experimental results show that proposed components we add can enhance the segmentation performance signally. To further assess the importance of our method, we conduct ablation experiments and corroborate our purposed method in both quantitative analysis and visual comparison.
\subsubsection{Quantitative Results}

Tab. \ref{tab1} lists the quantitative comparison between our model and baseline or other advanced networks. Compared with the baseline model U-Net, the Pixel Accuracy of our MFA-Net has the most obvious improvement, while Jaccard Index has the fewest improvement. In most cases, our model achieved the best segmentation performance, but was inferior to the AttUNet model on the LiTS 2017 dataset. Tab. \ref{tab2} shows results of ablation experiments on two datasets. Comparing along the rows, we observe a statistically increase in Jaccard Index and Pixel Accuracy with the addition of any SE block in comparison to the normal version of U-Net. Also, as shown in Tab. \ref{tab2}, the introduction of the SE module only in the spatial dimension yields a higher increase than that in the channel dimension only. The experimental results confirmed the effectiveness of the attention module SCSE we introduced.

\subsubsection{Qualitative Results:}Fig. \ref{fig4} presents results for liver tumor segmentation by different modules. We show the input scan, ground truth annotations and predicted segmentation by using baseline model and different strategies adding SE block. We observe the baseline U-Net has a poor performance when the contrast between the liver tumor and the surrounding tissue is relatively close, and the introduction of attention mechanism can improve the accuracy for these cases. Clearly, our model achieves close results with comparison models, and performed better in some images.

\section{Conclusion}
In this paper, we introduce MFA-Net, a network with improved SE blocks. Past studies in applying the SE block into the U-Net model often deployed it only in the channel dimension but ignored the spatial dimension. Therefore, we applied the SE module to both dimensions. We carried out extensive experiments to evaluate our proposed method. The analysis on two datasets demonstrates the effectiveness of our model. Several ablation experiments are also implemented to verify that the spatial excitation is more effective. In the future, we would like to investigate more models with the SE block and we believe it will become the core component of more human-centric applications.

\bibliography{MFA-Net}

\begin{thebibliography}{14}
\providecommand{\natexlab}[1]{#1}

\bibitem[{Iantsen et~al.(2021)Iantsen, Jaouen, Visvikis, and Hatt}]{iantsen2021squeeze}
Iantsen, A.; Jaouen, V.; Visvikis, D.; and Hatt, M. 2021.
\newblock Squeeze-and-excitation normalization for brain tumor segmentation.
\newblock In \emph{Brainlesion: Glioma, Multiple Sclerosis, Stroke and Traumatic Brain Injuries: 6th International Workshop, BrainLes 2020, Held in Conjunction with MICCAI 2020, Lima, Peru, October 4, 2020, Revised Selected Papers, Part II 6}, 366--373. Springer.

\bibitem[{Kamnitsas et~al.(2016)Kamnitsas, Ferrante, Parisot, Ledig, Nori, Criminisi, Rueckert, and Glocker}]{kamnitsas2016deepmedic}
Kamnitsas, K.; Ferrante, E.; Parisot, S.; Ledig, C.; Nori, A.~V.; Criminisi, A.; Rueckert, D.; and Glocker, B. 2016.
\newblock DeepMedic for brain tumor segmentation.
\newblock In \emph{Brainlesion: Glioma, Multiple Sclerosis, Stroke and Traumatic Brain Injuries: Second International Workshop, BrainLes 2016, with the Challenges on BRATS, ISLES and mTOP 2016, Held in Conjunction with MICCAI 2016, Athens, Greece, October 17, 2016, Revised Selected Papers 2}, 138--149. Springer.

\bibitem[{Liu, Deng, and Yang(2019)}]{liu2019recent}
Liu, X.; Deng, Z.; and Yang, Y. 2019.
\newblock Recent progress in semantic image segmentation.
\newblock \emph{Artificial Intelligence Review}, 52: 1089--1106.

\bibitem[{Long, Shelhamer, and Darrell(2015)}]{long2015fully}
Long, J.; Shelhamer, E.; and Darrell, T. 2015.
\newblock Fully convolutional networks for semantic segmentation.
\newblock In \emph{Proceedings of the IEEE conference on computer vision and pattern recognition}, 3431--3440.

\bibitem[{Milletari, Navab, and Ahmadi(2016)}]{milletari2016v}
Milletari, F.; Navab, N.; and Ahmadi, S.-A. 2016.
\newblock V-net: Fully convolutional neural networks for volumetric medical image segmentation.
\newblock In \emph{2016 fourth international conference on 3D vision (3DV)}, 565--571. Ieee.

\bibitem[{Nair and Hinton(2010)}]{nair2010rectified}
Nair, V.; and Hinton, G.~E. 2010.
\newblock Rectified linear units improve restricted boltzmann machines.
\newblock In \emph{Proceedings of the 27th international conference on machine learning (ICML-10)}, 807--814.

\bibitem[{Ramesh et~al.(2021)Ramesh, Kumar, Swapna, Datta, and Rajest}]{ramesh2021review}
Ramesh, K.; Kumar, G.~K.; Swapna, K.; Datta, D.; and Rajest, S.~S. 2021.
\newblock A review of medical image segmentation algorithms.
\newblock \emph{EAI Endorsed Transactions on Pervasive Health and Technology}, 7(27): e6--e6.

\bibitem[{Roy, Navab, and Wachinger(2018)}]{roy2018concurrent}
Roy, A.~G.; Navab, N.; and Wachinger, C. 2018.
\newblock Concurrent spatial and channel ‘squeeze \& excitation’in fully convolutional networks.
\newblock In \emph{Medical Image Computing and Computer Assisted Intervention--MICCAI 2018: 21st International Conference, Granada, Spain, September 16-20, 2018, Proceedings, Part I}, 421--429. Springer.

\bibitem[{Soler et~al.(2010)Soler, Hostettler, Agnus, Charnoz, Fasquel, Moreau, Osswald, Bouhadjar, and Marescaux}]{soler20103d}
Soler, L.; Hostettler, A.; Agnus, V.; Charnoz, A.; Fasquel, J.; Moreau, J.; Osswald, A.; Bouhadjar, M.; and Marescaux, J. 2010.
\newblock 3D image reconstruction for comparison of algorithm database: A patient specific anatomical and medical image database.
\newblock \emph{IRCAD, Strasbourg, France, Tech. Rep}, 1(1).

\bibitem[{Wen et~al.(2021)Wen, Li, Shen, Zheng, and Zheng}]{wen2021squeeze}
Wen, J.; Li, Z.; Shen, Z.; Zheng, Y.; and Zheng, S. 2021.
\newblock Squeeze-and-Excitation Encoder-Decoder Network for Kidney and Kidney Tumor Segmentation in CT Images.
\newblock In \emph{International Challenge on Kidney and Kidney Tumor Segmentation}, 71--79. Springer.

\bibitem[{Woo et~al.(2018)Woo, Park, Lee, and Kweon}]{woo2018cbam}
Woo, S.; Park, J.; Lee, J.-Y.; and Kweon, I.~S. 2018.
\newblock Cbam: Convolutional block attention module.
\newblock In \emph{Proceedings of the European conference on computer vision (ECCV)}, 3--19.

\bibitem[{Yuheng and Hao(2017)}]{yuheng2017image}
Yuheng, S.; and Hao, Y. 2017.
\newblock Image segmentation algorithms overview.
\newblock \emph{arXiv preprint arXiv:1707.02051}.

\bibitem[{Zeng et~al.(2019)Zeng, Xulei, Qiyun, Meng, and Le}]{zeng2019sese}
Zeng, Z.; Xulei, Y.; Qiyun, Y.; Meng, Y.; and Le, Z. 2019.
\newblock Sese-net: Self-supervised deep learning for segmentation.
\newblock \emph{Pattern Recognition Letters}, 128: 23--29.

\bibitem[{Zhang et~al.(2020)Zhang, Lan, Zeng, Jin, and Chen}]{zhang2020relation}
Zhang, Z.; Lan, C.; Zeng, W.; Jin, X.; and Chen, Z. 2020.
\newblock Relation-aware global attention for person re-identification.
\newblock In \emph{Proceedings of the ieee/cvf conference on computer vision and pattern recognition}, 3186--3195.

\end{thebibliography}

\end{document}